\documentclass[11pt]{article}
\pdfoutput=1
\usepackage{geometry}
\usepackage{coling2020}
\usepackage{times}
\usepackage{url}
\usepackage{latexsym}  
\usepackage{microtype}

\colingfinalcopy 

\usepackage{times}
\usepackage{latexsym}

\usepackage{datatool}
\usepackage{pdfpages} 

\usepackage[T1]{fontenc}
\usepackage[utf8]{inputenc}
\usepackage[frenchb]{babel}

\usepackage{varwidth}
\usepackage{array}

\DTLsetseparator{|} 

\usepackage{comment}
\usepackage{soul}
\usepackage{hyperref}

\title{NLP-CIC at SemEval-2020 Task 9$\mathpunct{:}$ Analysing sentiment in code-switching language using a simple deep-learning classifier}

  \author{ 
  Jason Angel \\
  CIC, Instituto Polit\'{e}cnico Nacional \\
  Mexico City, Mexico \\
  {\tt ajason08@gmail.com}
  \And
  Segun Taofeek Aroyehun \\
  CIC, Instituto Polit\'{e}cnico Nacional \\
  Mexico City, Mexico \\
  {\tt aroyehun.segun@gmail.com}
  \AND  
  Antonio Tamayo \\
  CIC, Instituto Polit\'{e}cnico Nacional \\
  Mexico City, Mexico \\
  {\tt ajtamayoh@gmail.com} 
  \And
  Alexander Gelbukh \\
  CIC, Instituto Polit\'{e}cnico Nacional \\
  Mexico City, Mexico \\
  {\tt www.gelbukh.com}
  } 
  
\date{}

\begin{document}
\maketitle

\begin{abstract}
Code-switching is a phenomenon in which two or more languages are used in the same message. Nowadays, it is quite common to find messages with languages mixed in social media. This phenomenon presents a challenge for sentiment analysis.
In this paper, we use a standard convolutional neural network model to predict the sentiment of tweets in a blend of Spanish and English languages. Our simple approach achieved a F1-score of $0.71$ on test set on the competition. We analyze our best model capabilities and perform error analysis to expose important difficulties for classifying sentiment in a code-switching setting.
\end{abstract}

\section{Introduction}
\label{intro}

\blfootnote{
    %
    %
    \hspace{-0.65cm}  
This work is licensed under a Creative Commons Attribution 4.0 International Licence. Licence details: http://creativecommons.org/licenses/by/4.0/.
}

 The phenomenon of combining two or more languages in the same message is known as code-switching or code-mixing \cite{gumperz1982discourse,myers1993common}.
Code-switching is an indicator of bilingual competence  \cite{Hamers1999BilingualityAB}, and it is also motivated by social and cultural factors such as social status, race,
age, etc. \cite{kim2006reasons}.
Although this phenomenon has been studied extensively in linguistics \cite{pfaff1979constraints,poplack1980sometimes,gumperz1982discourse,myers1993common,milroy1995introduction,lipski2005code,martinez2010spanglish,auer2013code}, it is still challenging for machines to process mixed natural languages. Code-switching is notoriously present on social media posts and chats such as Twitter, Facebook or WhatsApp;  consequently making it more difficult to process the sentiment expressed in such contents.


 In this work, we present a Convolutional Neural Network (CNN) system to predict the sentiment of a given code-mixed tweet. The sentiment labels are either positive, negative, or neutral, and the languages involved are English and Spanish. Our best model utilizes only Spanish word embeddings from tweets \cite{deriu2017leveraging} and does not require manual feature engineering. 




\section{Related work}

 Sentiment analysis is a widely studied task in monolingual, multilingual and cross-lingual settings. 
 For instance, monolingual opinion mining in a multilingual context~\cite{boiy2009machine},  multilingual sentiment analysis~\cite{balahur2014comparative}, and cross-lingual polarity detection~\cite{demirtas2013cross}. Language-independent approaches to sentiment analysis include the use of emoticons \cite{davies2011language} or emoticons and noisy labels \cite{narr2012language}.
 Sentiment analysis has not been extensively studied on code-switched content. A possible reason is the paucity of large annotated data covering several language pairs or combinations. The Sentimix shared task is an effort to address this problem on the Spanish-English and Hindi-English language pairs for sentiment analysis.
 
 Analysing opinions in tweets that blend Spanish and English is a difficult task \cite{vilares2017supervised}. Vilares et al. \shortcite{vilares2017supervised} obtained an accuracy of 59.34\% using lexical, syntactic, and N-gram features. They concluded that the task is challenging because of the presence of noise, difficulties with language identification and POS tagging, and the lack of annotated code-mixed lexicons and a large dataset.
 
 Another line of work related to code-switching is the development of contextual and static multilingual or cross-lingual text representations which cover multiple languages in the same vector space. Examples include LASER \cite{artetxe2019massively}, MUSE \cite{lample2017unsupervised}, and multilingual BERT\footnote{\url{https://github.com/google-research/bert/blob/a9ba4b8d7704c1ae18d1b28c56c0430d41407eb1/multilingual.md}}. These representations can be used to encode inputs for deep learning models. The effectiveness of these text representation approaches on code-switched texts remains an open question.

 


\section{Methodology}
 We describe the Spanglish dataset and our submitted models. 
\subsection{The Sentimix Spanglish dataset} \label{subsection_dataset}
 The Sentimix Spanglish dataset \cite{patwa2020sentimix} consists of a list of tweets blending English and Spanish text. The dataset is divided into train, development, and test sets holding $12002$, $2998$, and $3789$ samples for each set respectively. This dataset provides detailed annotations per word, which besides the English and Spanish tags includes named entities, in-word mixes, ambiguous and foreign words. A major part of the dataset is Spanish, being the mode language for $65.1\%$ of all tweets, whereas English is only $20.5\%$. We noticed that language statistics are proportionally distributed across the three partitions of the dataset. With regards to sentiment labels, Table \ref{table:dist_sentiment} presents the label distribution for training and development sets.


\begin{table}[htbp!]
\centering
\begin{tabular}{|c|c|c|c|}
 \hline
 Set & Sentiment & Samples & Proportion\\
 \hline
 & Negative\ (N)   &  2023   & 16.86\%  \\
 train &  Neutral  \ (T)    &  3974   & 33.11\%  \\
 & Positive \ (P)  &  6005   & 50.03\%  \\ \hline
 &Negative\ (N)   &  506    & 16.88\%  \\
 development    &Neutral  \ (T)   &  994    &  33.16\% \\
 &Positive \ (P)  &  1498   &  49.97\%  \\
 \hline
\end{tabular}
\caption{Distribution of sentiment labels in the dataset}
\label{table:dist_sentiment}
\end{table}
\subsection{Text normalization as pre-processing}
We normalize the input text using Ekphrasis \footnote{\href{https://github.com/cbaziotis/ekphrasis}{https://github.com/cbaziotis/ekphrasis}} to address the noise in social media text.
Specifically, the normalization process consists of:
\begin{itemize}
    \item Mapping URL, email, percent, money, phone, user, time, date, and numbers to a unique descriptive token.
    \item Labeling stylistic patterns such as uppercase, elongated, repeated, emphasized, and censored words. Examples of these can be seen in Table 3. 
    \item Word transformation suitable for social media content, such as word segmenter, spellchecker, and tokenizer.
\end{itemize}
 We apply the normalization to the entire dataset and noticed it had a marginal influence over the Spanish text. We did not apply normalization to the Spanish part  of the text.

\subsection{Neural Network architectures}
\paragraph{CNN model.}
 We used a standard architecture \cite{kim2014convolutional} with standard values as  hyperparameters, consisting of a single convolution layer with multiple filter sizes of 2, 3 and 4, each of them with 100 filters, followed by max-pooling and a dropout layer, to finally stack a fully connected layer which outputs the results. We used ReLU as activation function, Adam as optimizer and cross-entropy loss function as the optimization objective.The hyperparameters are: vocabulary size of 15000, batch size of 64, and dropout probability of 0.5. We predict using the best epoch result obtained out of 5 epochs.

Our embedding vectors were initialized using only 200 dimension Spanish word embeddings  \footnote{\href{https://www.spinningbytes.com/resources/wordembeddings/}{https://www.spinningbytes.com/resources/wordembeddings/}} which were trained on a collection of tweets \cite{deriu2017leveraging}. We did not use English embeddings because in previous experiments we noticed it does not contribute enough to the performance gain.
This is probably  due to the fact that Spanish is the majority language in the dataset, as we mentioned before. 

\paragraph{GRU model.} This submission used a bidirectional Gated Recurrent Unit (GRU)\footnote{We chose GRU over LSTM because the former has less parameters to learn.} over English-Spanish aligned word embeddings of dimension $300$ \cite{conneau2017word}. The hidden state of the GRU is of dimension 512. We used dropout with probability of $0.1$ on the outputs of the embedding and GRU layers. Also, layer normalization is applied on the hidden representation generated by the GRU layer. A fixed representation for a given tweet is derived by taking the average of the backward and forward hidden representation of the GRU layer. This serves as input to a dense layer with softmax activation function which output a probability distribution over the three labels. We used AdamW as optimizer to minimize the negative log-likelihood loss function for $10$ epochs. We use a batch size of $256$ and learning rate of $0.001$.

\section{Results and discussion}
 We obtained a precision score of $0.807$ and a recall score of $0.647$ on the competition, thus a $0.71$ F1 score was calculated for the competition rank\footnote{\href{https://competitions.codalab.org/competitions/20789\#learn\_the\_details-result}{https://competitions.codalab.org/competitions/20789\#learn\_the\_details-result} (Codalab user name (our team): ajason08)}.
Moreover, according to the competition guidelines, we also report class-wise F1 scores for test set, in which we achieved a lower score
\footnote{The Codalab scorer is using a less strict metric, probably a weighted scheme.}.
Under this metric, our aforementioned architecture achieved results shown at Table \ref{table:class-wise-results}. These results are similar to those obtained on our experiments on the development set, where we achieved a performance of {0.45} using macro-F1.

\begin{table}[htbp!]
\centering
\begin{tabular}{|c|c|c|c|c|c|c|c|}
 \hline
 &\multicolumn{3}{|c|}{CNN model} & \multicolumn{3}{c|}{GRU model}\\
 \cline{2-7}
  & Precision & Recall & F1-score & Precision & Recall & F1-score\\
 \hline
  Positive &    0.901   &   0.71    &   0.794	&    0.854   &   0.404    &   0.548 \\
  Negative &    0.547   &   0.375   &   0.445	&    0.155   &   0.341    &   0.213\\
  Neutral  &    0.082   &   0.403   &   0.136	&	 0.065   &   0.374    &   0.110\\
 \hline
\end{tabular}
\caption{Class-wise performance scores}
\label{table:class-wise-results}
\end{table}




\subsection{Error analysis} \label{posterioranalysis}

 Here an additional analysis is done to examine the results of our best model (the CNN model) via removing the text normalization stage and running cross-validation. Moreover, we perform error analysis to find common situations in which our system fails to make the correct prediction. These analysis were done using development set labels.

We found that removing the normalization step reduces the model performance in some points. Specifically, the macro-F1 score without the presence of Ekphrasis normalization is 0.42, hence a performance reduction of 3\%. 

 We present five categories that highlight some of the errors that our best model made.
 The analysis was made over a stratified sample of $300$ examples on the development set. Due to space limitations, we show a few examples of these categories along with the gold label (L) and our prediction (P) in Table 3. We consider the following categories:
\newpage
\includepdf[fitpaper=true, pages=-]{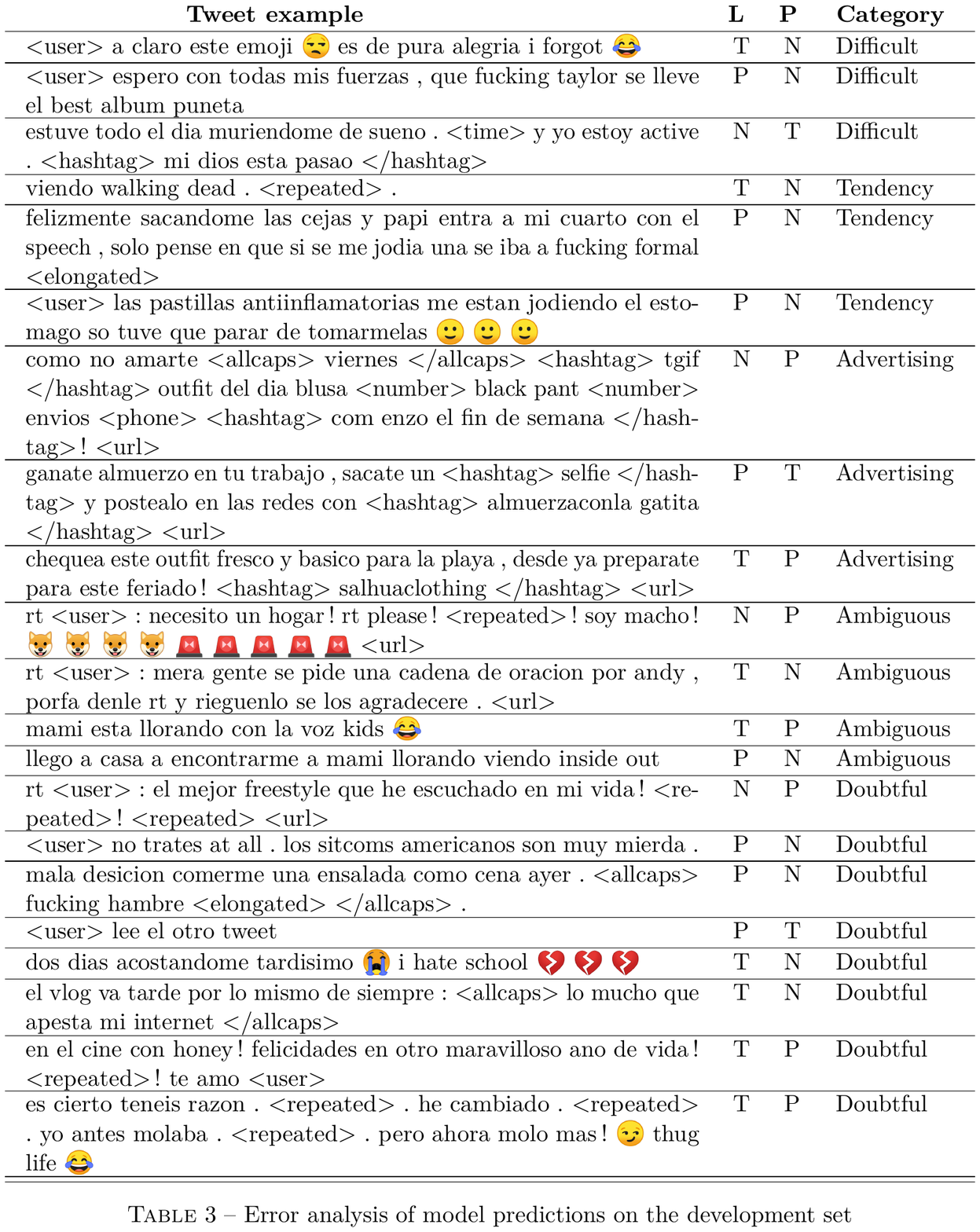}
 
 
\begin{itemize}
    \item \textbf{Difficult:} examples our best model classified wrongly and we consider hard because of informal style which leads to out-of-vocabulary issues, and the high level of common sense or pragmatics understanding probably involved. Also, we noticed that often the effect of code-switching is sarcastic.
    \item \textbf{Negative tendency:} the use of highly negative words (e.g. vulgar expressions) seems to bias the model towards predicting a negative sentiment, however, this only occurs in the English portion of the text. We notice it because some English words appears in the embedding vocabulary, probably from people doing Spanish-English code-switching.
    \item \textbf{Advertising:} These are tweets promoting some product or service. Surprisingly, most of them have
    neutral or negative label.    
    Conversely, our best model rates them mostly as conveying positive or neutral sentiment.
    \item \textbf{Ambiguous labels:} we hypothesize that the model gets confused when it finds samples with a very similar narrative but different labels. For example, the tweet ``\underline {mami esta llorando con la voz kids}"\footnote{"My mom is crying while watching 'La Voz Kids'"} is referring to a hilarious situation, and it receives a ``neutral" label, however a second tweet ``llego a casa a encontrarme a \underline {mami llorando viendo inside out}''\footnote{"I arrive at home and I found my mom crying while watching 'Inside Out'"} although is referring to a very similar situation, it receives a ``positive" label, we noticed that for both tweets the source of the sentiment comes from the relation between the adult and emotive show. We think these instances are a source of confusion for the model.
    
    \item \textbf{Doubtful labels:} some tweet labels can be related to subjectivity in the annotation process. 
    We consider that these examples are incorrectly labeled.
    
\end{itemize}


\section{Conclusions}
 Code-switching is an interesting problem holding an important presence in social media, which combined with informal writing style increases the challenges for social media processing such as sentiment analysis.
 We experimented with the Sentimix Spanglish dataset using CNN model and only Spanish embeddings. We achieve a precsion, recall, and F1 score of 0.80, 0.64, and 0.71 respectively.
 Our analyses suggest that a deep learning model can be easily biased by the presence of cue words such as vulgar expressions for sentiment analysis. We found that this occurs mostly when the cue word is in English. This observation requires a deeper analysis.
 We also highlight the need to address complex language usage such as informality and sarcasm.
  
 Furthermore, we also pointed out that subjectivity in the annotation of sentiment labels is a problem that deserves to be addressed.
 We plan to test contextual multilingual embeddings (e.g. Bert) and leverage the language tags and other non-linguistic constructs such as hashtags and emojis.




\bibliographystyle{coling}
\bibliography{references}

\end{document}